\documentclass[runningheads]{llncs}
\usepackage[T1]{fontenc}
\usepackage{graphicx}
\usepackage{booktabs}
\usepackage{amsmath}
\usepackage{url}
\usepackage{float}

\begin{document}

\title{Synthetic Augmentation for Data-Scarce Object Detection: Allocation, Dose, and Filtering Across Six Datasets}
\titlerunning{Synthetic Augmentation for Data-Scarce Detection}

\author{Jo\~ao Guedes F\'elix\inst{1} \and Atabak Dehban\inst{1} \and Rodrigo Ventura\inst{1}}
\authorrunning{J. G. F\'elix, A. Dehban, and R. Ventura}
\institute{Institute for Systems and Robotics, University of Lisbon, Portugal}

\maketitle

\begin{abstract}
Robotic perception for capturing uncooperative orbital objects and inspecting variable industrial products must often operate with limited labelled imagery. We examine a modular synthetic-augmentation pipeline, evaluating generation planning and post-generation filtering as separate interventions, with computational cost as a design constraint. Across six object-detection datasets, FLUX.1-Fill with class-specific LoRA adaptation produced dataset-dependent dose effects; the validation-selected augmentation improved mAP@50:95 on four datasets and reduced it on two. Filtering did not yield consistent gains, and FID, KID, precision, recall, and LPIPS did not consistently track detector changes. The results do not establish that image quality alone identifies useful samples. They motivate standardised evaluation and inexpensive candidate-utility measures that could identify useful samples before detector retraining, reducing repeated training cost.
\end{abstract}

\section{Introduction}

Robotic capture of uncooperative orbital objects illustrates a setting in which acquiring representative real imagery is difficult. Inspection and repair in circular manufacturing face a related data constraint when products vary in appearance and condition. These applications motivate a broader question: can object detection improve when only a small labelled real dataset is available? Synthetic augmentation may help, but plausible images need not add useful training information. The present experiments test the approach on general detection benchmarks rather than on these application domains.

The value of synthetic data may depend on what is generated, how much is added, and which candidates are retained. We study these choices in a modular, diffusion-based pipeline, comparing generation-planning approaches, including one with detector feedback, and assessing post-generation filtering separately. Computational cost is a design constraint.

We also test whether the effect changes with synthetic-data quantity and whether image-level metrics track detector performance.

\section{Related Work}

Diffusion-based approaches have been used to generate object-detection training data. ODGEN combines object- and layout-conditioned synthesis for domain-specific detection~\cite{zhu2024odgendomainspecificobjectdetection}; FLORA studies efficient data generation through LoRA adaptation of FLUX~\cite{patricio2025floraefficientsyntheticdata}. Controllable diffusion augmentation has also used semantic scores to reject unsuitable candidates~\cite{fang2024controllable}. Earlier detector studies found that rendering choices affect how well synthetic-image training transfers to real scenes~\cite{rozantsev2015rendering}. Flow matching provides a continuous-time approach to generative modelling by learning vector fields along paths between noise and data distributions~\cite{lipman2023flowmatching}. These works motivate synthetic generation and candidate assessment, but results from different datasets and protocols do not provide a controlled ranking for the present study.

Dataset allocation and selection offer a separate line of work. SD4OD analyses dataset-level deficits to guide synthetic-data creation~\cite{clearbox_sd4od}. Coreset selection for object detection formalises representativeness and diversity in subset construction~\cite{lee2024coreset}. GenDataAgent uses downstream-model feedback and gradient-based signals to guide image-classification augmentation~\cite{li2025gendataagent}; a similar feedback strategy could be explored for object detection. Our study compares a dataset-statistics allocation rule with one detector-weighted variant, then evaluates a separate post-generation selector.

Image-level metrics capture different properties of generated data. FID and KID compare feature distributions~\cite{heusel2017fid,binkowski2018demystifying}; generative precision and recall estimate feature-space fidelity and coverage~\cite{kynkaanniemi2019improvedprecisionrecall}; LPIPS measures perceptual differences between paired images~\cite{zhang2018lpips}. Analysing these measures alongside detector metrics makes it possible to examine how image-level assessments relate to downstream detection performance.

\section{Method}

\subsection{Generation and Allocation}

\begin{figure}[t]
\centering
\includegraphics[width=\textwidth]{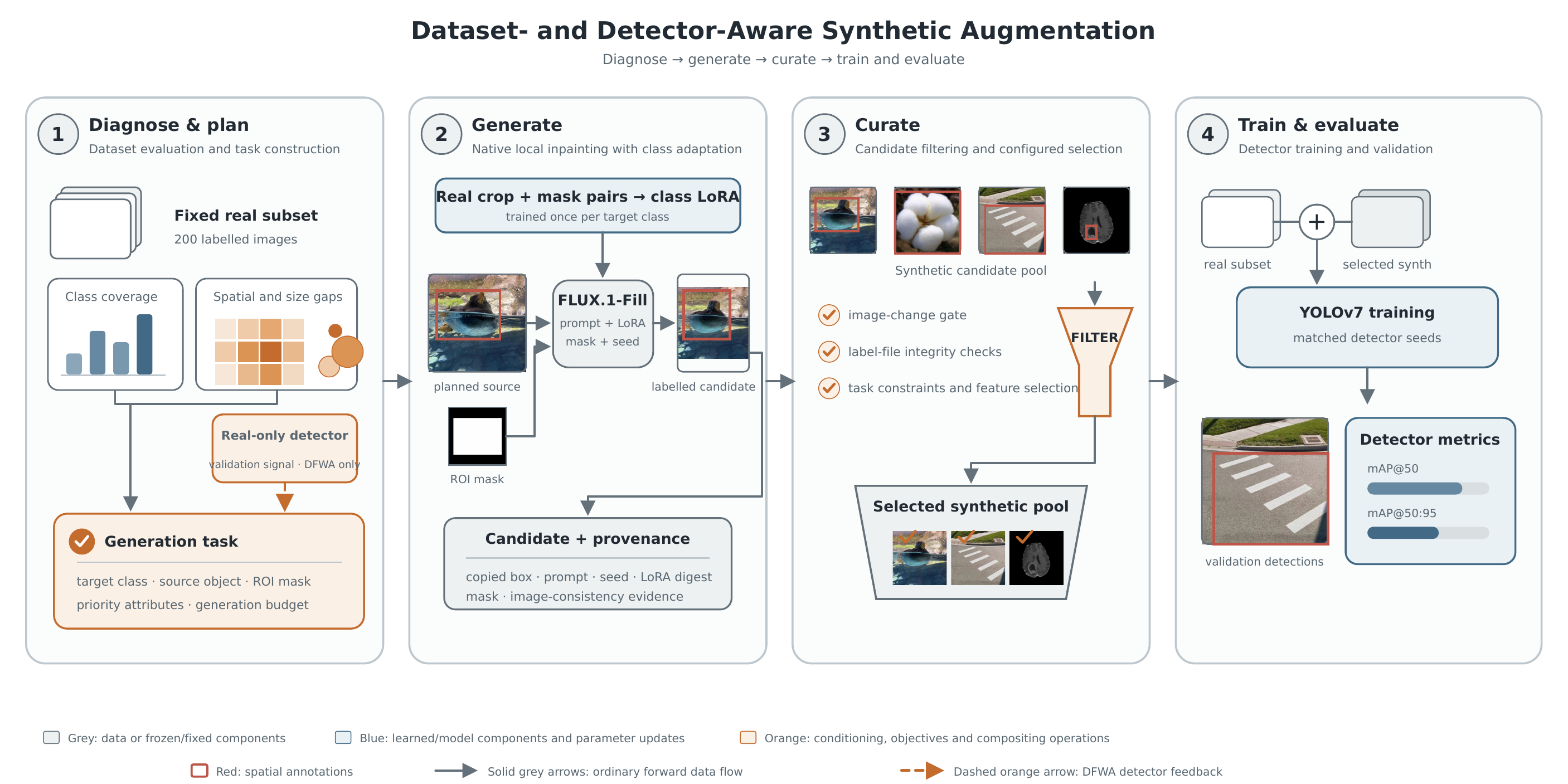}
\caption{Experimental pipeline from dataset analysis and generation planning through LoRA adaptation, local inpainting, candidate selection, and detector evaluation. The detector signal is used only by DFWA to set pre-generation priorities.}
\label{fig:pipeline}
\end{figure}

The pipeline analyses real training annotations, allocates generation requests, adapts class-specific LoRA modules, and inpaints selected objects in source scenes. Figure~\ref{fig:pipeline} summarises the stages. For Detector-Feedback-Weighted Allocation (DFWA), validation results from the real-only detector set generation priorities before candidates are produced.

The four allocation strategies vary in the information used to choose requests. Random Allocation selects source objects without target-group priorities. Class-Quota Allocation assigns quotas at class level. Dataset-Statistics-Guided Allocation (DSGA) directs requests towards under-represented class-conditioned buckets based on object size, aspect ratio, and position. DFWA multiplies the DSGA gap score for target group $t$ by a class-level weakness estimate:
\begin{equation}
P_t = B_t W_c, \qquad W_c=0.5(1-\mathrm{AP}_c)+0.5C_c,
\end{equation}
where $B_t$ is the dataset-statistics gap score, $\mathrm{AP}_c$ is validation AP@50:95, and $C_c$ is the normalised validation confusion rate for class $c$. The detector weights remain unchanged during this allocation stage.

\begin{figure}[t]
\centering
\includegraphics[width=\textwidth]{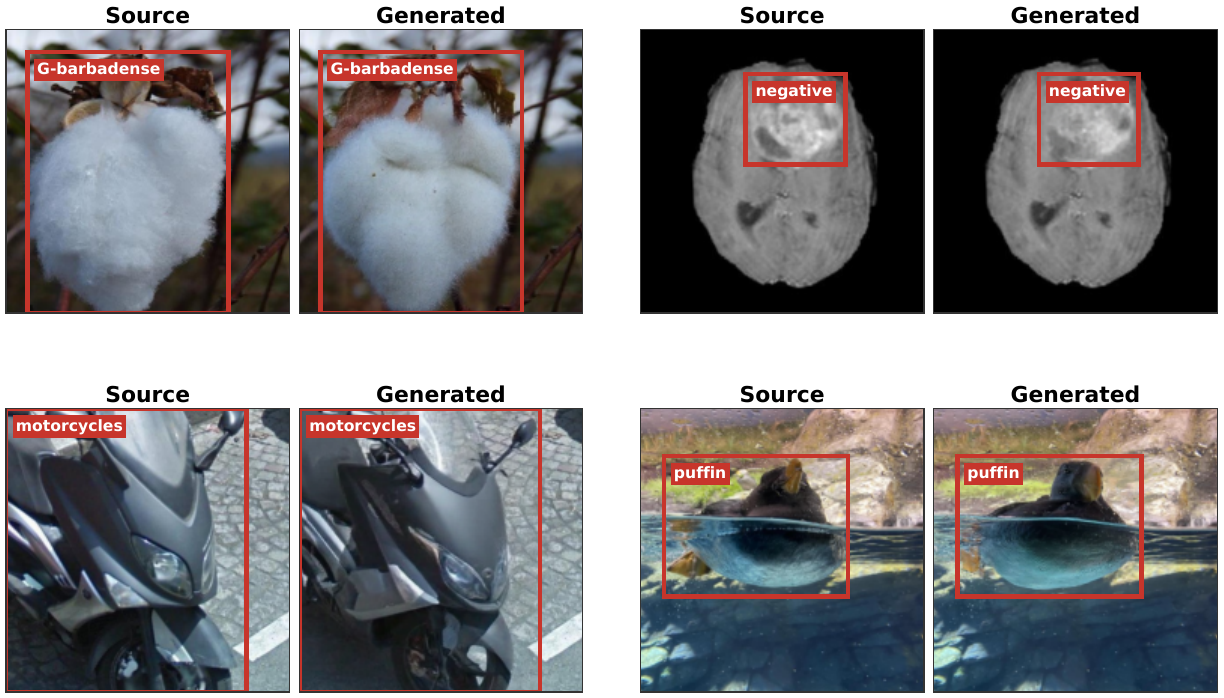}
\caption{Representative source and generated pairs from four datasets. Red boxes and class tags show the inherited YOLO annotations. Generated panels are inpainted modifications of the source images.}
\label{fig:generation_examples}
\end{figure}

For each dataset, a deterministic rule selects 200 labelled real training images. Synthetic conditions add $K\in\{50,100,200\}$ images from the selected pool, giving training sets denoted $200+K$. FLUX.1-Fill is adapted with class-specific LoRA modules using annotation-derived object crops and masks, except for MRI, where full-frame examples with class-specific masks are used. Each generation request specifies the source object, mask, class prompt, and sampling seed; the generated region is composited into its source scene, retaining the inherited detection annotations. LoRA uses rank $r=32$ and scaling $\alpha=32$~\cite{hu2021loralowrankadaptationlarge}. The experiment varies allocation and dose, not the pretrained diffusion architecture. Figure~\ref{fig:generation_examples} shows representative source and generated images with their inherited annotations. Source-image attribution and licensing details are listed in Ref.~\cite{source_image_credits}.

\subsection{Filtering}

Filtering addresses candidate choice after generation and is evaluated only on candidates generated by DSGA. Integrity gates check whether the edit remains local, whether the mask boundary has a pronounced discontinuity, and whether enough pixels within the mask changed. The minimum-change ratio $r_i$ is the proportion of masked pixels with at least one RGB channel changing by more than five intensity levels; candidates require $r_i\geq0.15$.

The selector maintains class and target-group counts and limits reuse of a source image. When a scene contains multiple annotations, all inherited labels contribute to class-count constraints; feature ranking represents the edited target object from its inherited box using a frozen real-only YOLOv7 detector. L2-normalised features are compared by cosine distance to same-class real objects and to already selected candidates. Feature distance ranks candidates but does not determine whether a candidate is accepted or rejected. If passing candidates cannot fill a quota within the source-reuse limit, candidates from the corresponding unfiltered subset are used as fallbacks.

\subsection{Evaluation Protocol}

We evaluate six datasets from Roboflow-100~\cite{ciaglia2022roboflow100}: Aquarium, Cotton, MRI, Road Traffic, Robotmaster, and Underwater. YOLOv7~\cite{wang2022yolov7} is trained on each real-only and augmented condition using the same three detector seeds (11, 23, and 37); paired changes are calculated relative to the real-only run with the matching seed. The primary metric is mAP@50:95. Allocation and dose selection use validation performance. After selection, the unfiltered DSGA dose chosen for each dataset is evaluated on the held-out test partition. The fixed training schedule is 100 epochs.

Image metrics are calculated separately from detector evaluation. For the association analysis, FID, KID, precision, recall, and LPIPS are measured on each unfiltered DSGA dose subset. FID and KID use both the complete dataset partitions and the 200-image generation-source subset as references. Precision and recall use validation and test images as an image-only reference; no test labels or detector outputs enter these calculations. LPIPS is averaged over paired source--generation images. For each dataset, we subtract the mean across its three doses from each metric value and paired validation mAP change. Pearson correlation is then calculated over the 18 dataset-dose pairs. This measures whether dose-level changes in an image metric accompany dose-level changes in detector performance within datasets, rather than comparing absolute metric values across datasets. The dose subsets are nested, so the 18 pairs are not independent observations.

\section{Results}

\subsection{Allocation Strategy and Synthetic Dose}

Table~\ref{tab:strategy} shows a trade-off across the 18 dataset-dose comparisons. Random Allocation produced the largest equal-weight mean paired change, while DSGA had positive mean changes in 12 comparisons, one more than Random Allocation. DSGA was carried forward because the validation selection criterion prioritised the frequency of positive changes. For Robotmaster, DSGA and DFWA specified identical target groups and therefore produced the same results.

\begin{table}[H]
\caption{Validation allocation summary across available matched dataset-dose comparisons. Mean $\Delta$ is the equal-weight average of the 18 dataset-dose mean changes; positive cells count means above the real-only baseline.}
\label{tab:strategy}
\centering
\small
\begin{tabular}{lrr}
\toprule
Strategy & Mean $\Delta$ & Positive cells \\
\midrule
Random Allocation & $\mathbf{+0.0291}$ & 11/18 \\
DSGA & $+0.0237$ & \textbf{12/18} \\
Class-Quota Allocation & $+0.0102$ & 11/18 \\
DFWA & $+0.0098$ & 9/18 \\
\bottomrule
\end{tabular}
\end{table}

Table~\ref{tab:dose} shows non-monotonic dose responses on all six datasets. The best of the tested doses exceeded the real-only mean in five datasets, with gains ranging from 0.009 mAP@50:95 for Aquarium to 0.179 for MRI. Every tested Road Traffic dose remained below its real-only baseline. The dose with the largest mean change varied across datasets. Doses above 200 were not tested.

\begin{table}[H]
\caption{Paired validation $\Delta$ mAP@50:95 for DSGA relative to the same-seed real-only result. Entries are means $\pm\sigma_{\Delta}$ over three paired seed differences.}
\label{tab:dose}
\centering
\scriptsize
\setlength{\tabcolsep}{4pt}
\begin{tabular}{lrrr}
\toprule
Dataset & $K=50$ & $K=100$ & $K=200$ \\
\midrule
Aquarium & $\mathbf{+0.0090\pm0.0152}$ & $-0.0091\pm0.0136$ & $-0.0090\pm0.0111$ \\
Cotton & $\mathbf{+0.0188\pm0.0592}$ & $+0.0079\pm0.0306$ & $+0.0113\pm0.0279$ \\
MRI & $\mathbf{+0.1787\pm0.1477}$ & $+0.0781\pm0.0367$ & $+0.1470\pm0.1406$ \\
Road Traffic & $-0.0202\pm0.0199$ & $\mathbf{-0.0145\pm0.0098}$ & $-0.0606\pm0.0098$ \\
Robotmaster & $-0.0740\pm0.3389$ & $\mathbf{+0.0349\pm0.3659}$ & $+0.0229\pm0.3405$ \\
Underwater & $+0.0275\pm0.0697$ & $+0.0214\pm0.0219$ & $\mathbf{+0.0563\pm0.0365}$ \\
\bottomrule
\end{tabular}
\end{table}

\subsection{Filtering and Image Metrics}

Filtering increased mean mAP@50:95 in 7 of 18 matched dataset-dose comparisons and decreased it in the other 11; filtered conditions exceeded real-only training in 9 comparisons. The direction varied by dataset. In Robotmaster, the integrity gates and source-reuse limit left too few passing candidates to fill 95 of 350 requested positions across doses, so the selector used examples from the corresponding unfiltered subsets for those slots. This is a limitation of the evaluated selector under that candidate pool.

Table~\ref{tab:correlation} reports the associations between dose-level image metrics and detector changes. The largest positive value was $r=0.32$ for train-200 KID, and recall had the most negative value ($r=-0.41$). These are exploratory associations across 18 observations, not evidence of predictive utility.

\begin{table}[H]
\centering
\caption{Within-dataset-centred Pearson correlations between image metrics and paired validation $\Delta$mAP@50:95 ($n=18$ dataset-dose observations). Dose subsets overlap; coefficients are descriptive and no significance test was conducted.}
\label{tab:correlation}
\small
\setlength{\tabcolsep}{6pt}
\begin{tabular}{p{7.4cm}r}
\toprule
Image metric and reference & Within-dataset $r$ \\
\midrule
FID, full train--validation--test reference & 0.12 \\
FID, train-200 reference & 0.12 \\
KID, full train--validation--test reference & 0.21 \\
KID, train-200 reference & 0.32 \\
LPIPS, source--generation pairs & $-0.05$ \\
Generative precision, validation/test reference & $-0.003$ \\
Generative recall, validation/test reference & $-0.41$ \\
\bottomrule
\end{tabular}
\end{table}

\subsection{Held-Out Test Evaluation}

The validation-selected unfiltered DSGA condition improved test mAP@50:95 in four datasets and decreased it in two (Table~\ref{tab:test}). The equal-weight mean paired change across the six datasets was $+3.02$ percentage points, with the largest absolute increase for MRI ($+10.18$ points) and the largest decrease for Road Traffic ($-2.79$ points). Relative to each dataset's real-only mean, the changes ranged from $-6.2\%$ for Road Traffic to $+87.7\%$ for Underwater; the equal-weight mean of the six dataset-specific percentage changes was $+29.8\%$.

\begin{table}[H]
\caption{Test mAP@50:95 for this study and published RF7 YOLOv7 results~\cite{zhu2024odgendomainspecificobjectdetection,patricio2025floraefficientsyntheticdata}. This study reports held-out means $\pm\sigma_{\mathrm{seed}}$ in percent; gains in parentheses are changes from the real-only baseline in percentage points.}
\label{tab:test}
\centering
\scriptsize
\setlength{\tabcolsep}{1pt}
\begin{tabular*}{\textwidth}{@{\extracolsep{\fill}}lccrr@{}}
\toprule
Dataset ($K$) & Real-only & DSGA & ODGEN ($200+5000$) & FLORA ($200+500$) \\
\midrule
Aquarium (50) & $35.93\pm1.81$ & $35.50\pm0.90\,(-0.43)$ & $29.6\to38.5\,(+8.9)$ & $29.6\to41.0\,(+11.4)$ \\
Cotton (50) & $8.14\pm2.48$ & $\mathbf{10.97\pm1.44}\,(+2.83)$ & $20.5\to43.2\,(+22.7)$ & $20.5\to44.5\,(+24.0)$ \\
MRI (50) & $21.40\pm9.41$ & $\mathbf{31.58\pm1.58}\,(+10.18)$ & $27.4\to41.5\,(+14.1)$ & $27.4\to37.6\,(+10.2)$ \\
Road Traffic (100) & $\mathbf{44.77\pm1.84}$ & $41.98\pm0.33\,(-2.79)$ & $41.0\to43.8\,(+2.8)$ & $41.0\to62.3\,(+21.3)$ \\
Robotmaster (100) & $18.58\pm16.62$ & $\mathbf{21.64\pm18.90}\,(+3.06)$ & $26.5\to34.7\,(+8.2)$ & $26.5\to38.2\,(+11.7)$ \\
Underwater (200) & $6.02\pm2.24$ & $\mathbf{11.30\pm4.79}\,(+5.28)$ & $19.4\to22.0\,(+2.6)$ & $19.4\to24.5\,(+5.1)$ \\
\bottomrule
\end{tabular*}
\end{table}

The ODGEN and FLORA values are taken from their papers; neither pipeline was run here. They are contextual references: despite both reporting RF7 results, their budgets and evaluation protocols differ, and no shared end-to-end benchmark or reporting format currently supports exact reproduction across methods.

\section{Discussion and Conclusion}

Adding generated images did not produce a uniform increase in detector performance. Dose responses differed by dataset, and the held-out results do not support a general claim of improvement. Stronger gains on specialised datasets are consistent with greater marginal value where pretrained representations may be weaker; the experiments do not isolate this factor from other dataset differences. Performance in circular manufacturing remains untested.

Filtering did not consistently improve detection. The image metrics describe distributional or perceptual properties, and their measured associations did not reliably indicate detector changes. Candidate-level utility estimates could guide generation towards samples likely to help a target detector, reducing repeated generation and retraining. Detector--generator feedback is one possible route, although it would increase computational cost and require controls that separate generator changes from detector adaptation.

Three detector seeds and one synthetic pool per condition reveal seed variation, not full-pipeline generalisation. Future benchmarks should share splits, generation budgets, and detector protocols, and include independent synthetic pools and detector runs.

Code and reproduction resources will be added at \url{https://github.com/JGuedesFelix/flux-od}.

\bibliographystyle{splncs04}
\bibliography{references}

\end{document}